\documentclass{article}

\usepackage[english]{babel}
\usepackage[T1]{fontenc}
\usepackage{amssymb}
\usepackage[utf8]{inputenc}
\usepackage{amsmath}
\usepackage[numbers]{natbib}
\usepackage[final]{nips_2017}
\usepackage[utf8]{inputenc} 
\usepackage[T1]{fontenc}    
\usepackage{hyperref}       
\usepackage{url}            
\usepackage{booktabs}       
\usepackage{amsfonts}       
\usepackage{nicefrac}       
\usepackage{microtype}      

\title{Riemannian tangent space mapping and \\elastic net regularization for cost-effective EEG markers of brain atrophy in Alzheimer's disease}

\author{
  Wolfgang Fruehwirt \\
  Medical University of Vienna  \&   \\
  University of Oxford
  \And
Matthias Gerstgrasser \\
University of Oxford \\
  \AND
Pengfei Zhang \\
University of Oxford \\
  \And
Leonard Weydemann\\
Medical University of Vienna\\
\And
Markus Waser \\
Technical University of Denmark \\
  \And
Reinhold Schmidt \\
Medical University of Graz \\
  \And
Thomas Benke \\
Medical University of Innsbruck \\
  \And
Peter Dal-Bianco \\
Medical University of Vienna \\
  \And
Gerhard Ransmayr \\
Linz General Hospital \\
  \And
Dieter Grossegger \\
Dr. Grossegger \& Drbal GmbH \\
  \And
Heinrich Garn  \\
Austrian Institute of Technology\\
  \And
Gareth W. Peters  \\
University College London\\
  \And
Stephen Roberts  \\
University of Oxford\\
  \And
Georg Dorffner   \\
Medical University of Vienna\\
}

\begin{document}

\maketitle

\begin{abstract}
The diagnosis of Alzheimer's disease (AD) in routine clinical practice is most commonly based on subjective clinical interpretations. Quantitative electroencephalography (QEEG) measures have been shown to reflect neurodegenerative processes in AD and might qualify as affordable and thereby widely available markers to facilitate the objectivization of AD assessment. Here, we present a novel framework combining
Riemannian tangent space mapping and elastic net regression for the development of brain atrophy markers. While most AD QEEG studies are based on small sample sizes and psychological test scores as outcome measures, here we train and test our models using data of one of the largest prospective EEG AD trials ever conducted, including MRI biomarkers of brain atrophy. 
\end{abstract}
\section{Introduction}
Having been successfully applied in domains such as computer vision 
\cite{Tuzel2007}, radar signal processing \cite{Barbaresco2008}, and diffusion tensor imaging \cite{Pennec2006} for years, the introduction of a Riemannian manifold of symmetric positive-definite (SPD) matrices to brain signal 
analysis represents a powerful alternative to more traditional information extraction protocols.
Only recently has it been shown that Riemannian Brain-Computer Interface (BCI) methods outperform 
state-of-the-art Euclidian spatial filtering and machine learning techniques \cite{Congedo2017}.
Five recent international BCI competitions -- including last year’s Microsoft Cortana brain decoding challenge -- have been won using Riemannian geometry \cite{Congedo2017, Barachant2016}.\\ 
Several reasons for this success have been proposed in the literature. First, in the form of covariance matrices, SPD matrices are understood to be excellent representations of the raw electrophysiological brain signal, while reducing its unwanted variations \cite{Kalunga2016}. They have therefore become fundamental elements in methods such as common spatial pattern and canonical correlation analysis.
Second, SPD matrices are traditionally treated within Euclidian frameworks, ignoring their intrinsic non-Euclidian structure. Neglecting this fundamental characteristic may lead to deficient results \cite{Arsigny2007}. \\
These points not only have been found advantageous in BCI design but also make a strong case for the use of a Riemannian SPD matrix manifold in the assessment of neuronal degeneration as can be found in Alzheimer’s disease (AD). \\
AD is the most common form of dementia and ultimately fatal.
The combination of its severity and looming global epidemic scale -- caused by the ageing of our society -- makes AD a major public health concern \cite{AlzheimerAssoc2014}. Due to its degenerative nature, early accurate diagnosis and effective clinical monitoring are crucial. However, when it comes to routine clinical practice, AD assessment is most commonly done by subjective clinical interpretations at an already progressed stage of the disease.
So far, no cost-effective, widely-used biomarkers have been established to facilitate the objectivization of diagnosis and disease progression assessment. To promote the screening and monitoring of as many individuals as possible, such markers should not be dependent on costly equipment, such as MRI, or PET scanners. Therefore, we focus on inexpensive apparatuses, namely electroencephalography (EEG) devices. Their non-invasiveness and low noise level adds to their suitability for large-scale use in irritable patients such as those found within the spectrum of AD. Additionally, research suggests that quantitative electroencephalography (QEEG) reflects neurodegenerative processes in AD (for reviews, see \cite{Vecchio2013,Drago2011,Dauwels2011}).\\
Therefore, we aim to develop a Riemannian framework for QEEG markers of neuronal degeneration in AD and empirically investigate its usefulness. To be able to combine the merits of Riemannian geometry with the advantages of sophisticated Euclidean regularization and variable selection techniques like the elastic net (see \ref{elasticnet}), we map SPD matrices into the tangent space (see \ref{features}).
Sustaining the distance relationship of elements, this projection and a subsequent vectorization allows to treat SPD matrices as Euclidean entities.\\
All existing Riemannian brain signal analysis methods use covariance as measure of dependence, thereby implicitly assuming a multivariate Gaussian distribution of data and linear associations between the activities of brain regions. 
However, both properties might not be fulfilled \cite{Tong2009, Waser2016}.
Hence, we examine the usefulness of rank correlation for constructing SPD matrices -- capturing non-linear relationships in data that is often far from normally distributed.
For model training and testing, we use one of the largest AD EEG data sets ever collected in a prospective manner, including MRI biomarkers of brain atrophy.
As frequency-specific QEEG information has been proven useful in the AD domain  \cite{Waser2016}, we furthermore analyze the effectiveness of a special type of spatiofrequential SPD matrix.
Finally, to evaluate the real added value of tangent space mapping, we compare results achieved by this method with those achieved by using regular Euclidean procedures.
\section{Materials and Methods}
\subsection{Experimental data}
AD patients were prospectively recruited at four tertiary memory clinics (Medical Universities of Graz, Innsbruck, Vienna, and the General Hospital Linz, PRODEM cohort study by the Austrian Alzheimer Society \cite{Seiler2012}, supported by the Austrian Research Promotion Agency FFG, project no. 827462).
We exclusively analyzed participants (N = 110) who had a structural MRI scan within 60 days of the baseline EEG measurement, a maximal MMSE score \cite{Folstein1975} of 28, and a CDR \cite{Hughes1982} from 0.5 to 1.\\
Acquisition of structural T1-weighted images was accomplished on 1.5 and 3 Tesla 
MR scanners (Siemens). We used FreeSurfer volumetric analyses \cite{Fischl2012} to build two MRI biomarkers of brain atrophy, i.e., cerebral volume and hippocampal volume divided by the total intracranial volume (ratios referred to as BrainVol and HippVol). \\
Continuous EEG (alpha trace EEG recorder, 10-20 electrode placement) was analyzed for an eyes-closed resting condition (EC, 180 sec; prediction of BrainVol) and the encoding period of a paired-associate word list task (WLT, adapted version of \cite{Plihal1997}, 140 sec; prediction of HippVol). Research has repeatedly shown (i) the importance of hippocampal activity during the WLT \cite{Cameron2001, Clark2017}, and (ii) the sensitivity of paired-associative memory to early AD-related changes \cite{Fowler2002, Pike2008}.\\
For details on the entire PRODEM experimental protocol and preprocessing pipeline, see \cite{Waser2016, Garn2014, Fruehwirt2016}.
\subsection{Feature generation}
\label{features}
To optimize the number of time points available, we determined the maximal signal length with guaranteed quasi-stationary properties using an augmented Dickey-Fuller test \cite{Dickey1979}. The de-artifacted signal was partitioned into 4-sec segments accordingly. SPD matrices were estimated using sample covariance (SCM, see \ref{covmat}) and Kendall rank correlation (KEN, see \cite{Kendall1938}).
In the following all procedures will be described by taking the example of sample covariance. \\
EEG segments were considered as $n$ by $t$ matrices $X_{s} \in \mathbb{R}^{nxt}, (s = 1 \ldots S)$, $n$ being the number of electrodes, $t$ being the number of time samples.
For each segment $s$ a spatial covariance matrix $C_{s}$ was estimated: 
\begin{equation}
\label{covmat}
C_{s} = \frac{1}{t-1} X_{s}X_{s}^{\intercal}
\in \mathbb{R}^{n \times n}
\end{equation}
To take distinct frequential aspects into account, we created a special type of matrix by band-pass filtering $X_{s}$ multiple times ($\delta$ = 2 to 4 Hz, $\theta$ = 4  to 8 Hz $\alpha$ = 8 to 13 Hz, $\beta_{1}$ = 13 to 15 Hz) and vertically concatenating the resulting signal to ${X}_s^{SF} = [{{X}}_{(\delta)}; {{X}}_{(\theta)}; {{X}}_{(\alpha)};{{X}}_{(\beta_{1})}]^{\intercal} \in \mathbb{R}^{4n\times t}$.
Then, the sample covariance matrix $C_{s}^{SF}$ was estimated:
\begin{equation}
C_{s}^{SF} = \frac{1}{t-1} X_{s}^{SF}X_{s}^{SF^\intercal} \in \mathbb{R}^{4n \times 4n}
\end{equation}
The common Euclidean distance ($d_{euc}$) between two matrices $C_1,C_2$ and the corresponding mean ($M_{euc}$) of several matrices $C_1,\ldots,C_N$ can be defined as
\begin{equation}
d_{euc}(C_{1},C_{2})=\lVert C_1 - C_2 \lVert_{F}
\end{equation}
\begin{equation}
\label{meuclid}
M_{euc}\left(C_{1} \ldots C_{N}\right) = \frac{1}{N} \sum_{i=1}^{N} C_{i}
\end{equation}
where $\lVert . \lVert_{F}$ denotes the Frobenius norm. 
However, the Euclidean space suffers from several disadvantages, as -- for instance -- the averaging of SPD matrices may lead to a \textit{swelling effect} (the determinant of the Euclidean mean can be strictly larger than the original determinants \cite{Arsigny2007}). To avoid such artifacts from geometry, a more natural metric for SPD matrices, the Log-Euclidean distance $d_{log}$, with the corresponding mean $M_{log}$, can be used (e.g., \cite{Yger2015}): 
\begin{equation}
d_{log}(C_{1},C_{2})=\lVert \log(C_1) - \log(C_2) \lVert_{F}
\end{equation}
\begin{equation}
\label{mlog}
M_{log}\left(C_{1} \ldots C_{N}\right) = \exp \left(\frac{1}{N} \sum_{i=1}^{N} \log\left(C_{i}\right)\right)
\end{equation}
Further, SPD matrices can be treated in their native Riemannian space using geodesic distance $d_{rie}$ and the Riemannian geometric mean, often referred to as Karcher mean \cite{Karcher1977}, $M_{rie}$, which minimizes the sum of squared $d_{rie}$:
\begin{equation}
d_{rie}(C_{1},C_{2}) = \lVert \textrm{log} \left( C_{1}^{-\frac{1}{2}} C_{2} C_{1}^{-\frac{1}{2}} \right) \lVert_{F} = \left[ \sum_{i=1}^{N} \textrm{log}^{2} \lambda_{i} \right]^{\frac{1}{2}} 
\end{equation}
\begin{equation}
\label{mrie}
M_{rie}\left(C_{1} \ldots C_{N}\right) = \textrm{argmin}_{C} \sum_{i=1}^{N} {d_{rie}}^2\left(C_{i},C\right)
\end{equation}
where $\lambda_{i}$ are the eigenvalues of $ C_{1}^{-\frac{1}{2}} C_{2} C_{1}^{-\frac{1}{2}}$. As $M_{rie}$ has no closed-form solution for $N$ > 2 , we optimized it using the relaxed Richardson iteration \cite{Bini2013}.
For a review on the advantages of Riemannian geometry in brain signal processing and detailed formal definitions, see \cite{Congedo2017,Tuzel2007}. \\
Sets of $C_{s}$ and $C_{s}^{SF}$ were separately averaged on patient level using the aforementioned mean calculation methods (\ref{meuclid}, \ref{mlog}, \ref{mrie}) for both EEG paradigms (EC and WLT), and both measures of dependence (COV and KEN), resulting in subject-specific matrices $M_{z}$ ($z$ being a patient) for all variants. For Riemannian (TAN$_{rie}$) and Log-Euclidean (TAN$_{log}$) tangent space-based features, $M_{z}$ of the corresponding mean type was mapped into the tangent space
\begin{equation}
F_{z} = \text{upper}\left(M_{G}^{-\frac{1}{2}} \text{Log}_{M_{G}}(M_{z})M_{G}^{-\frac{1}{2}}\right)\\
\end{equation}
where $M_{G}$ was computed alternatively using $M_{rie}$ (\ref{mrie})
for TAN$_{rie}$ or $M_{log}$ (\ref{mlog})
for TAN$_{log}$. 
For a formal definition of the Riemannian tangent space, see \cite{Barachant2012, Tuzel2007}. For the Euclidean control condition (EUC) both, the averaging on subject as well as group level, was done using $M_{euc}$ (\ref{meuclid}).\\
Applying upper(.) as an operator vectorizing the upper triangular part of a SPD matrix, the feature vectors $F_{z}$ of $C_{s}$ $\in$ $\mathbb{R}^{n(n+1)/2}$, given $n$ = 19 resulting in 190 dimensions, and $F_{z}$ of $C_{s}^{SF}$  $\in$ $\mathbb{R}^{4n(4n+1)/2}$, given $n$ = 19 resulting in 2926 dimensions, were created.  
\subsection{Elastic net regression and repeated nested cross-validation}
\label{elasticnet}
The elastic net \cite{Zou2005}
was used as a regularization and variable selection technique to estimates a sparse regression model based on $F_{z}$. It imposes a combination of the $\ell_{1}$ (lasso, \cite{Tibshirani1996}) and $\ell_{2}$ (ridge, \cite{Hoerl1970}) penalties on regression coefficients. While enjoying a similar sparsity of representation as the lasso, the elastic net encourages a grouping effect, where strongly correlated predictors -- as presumably present in our data set due to spatially adjacent electrode placement -- tend to be in or out of the model together \cite{Zou2005}.\\
We used a $10\times10$ two-level nested cross-validation 
to determine generalization performance. The inner loop was included to sensibly choose a value for the regularization parameter $\lambda$ with minimal expected generalization error \cite{Varma2006}. The $\lambda$ value that resulted in the lowest mean squared error (MSE) in the inner loop was used to fit models in the outer loop.
The parameter $\alpha$, representing the weight of lasso ($\ell_{1}$) versus ridge ($\ell_{2}$) optimization, was set at 0.5. Age and gender were introduced in BrainVol models, whereas the magnetic field strength (varying values of Tesla between centers might influence the analysis of smaller structures) was additionally introduced in the HippVol models. All variables were normalized before model fitting. To reduce the variability of prediction outcome resulting from random training–test set splitting, we repeated the entire nested cross-validation procedure $100$ times. This allowed us to average out variability and report the range of results of multiple permutations \cite{Shouten}.
\section{Results and Discussion}
Results are depicted in Table~\ref{results}. 
For both prediction problems (BrainVol, HippVol), the best models were of spatiofrequential nature (indicated in bold in the table), highlighting the importance of frequency-specific information for QEEG AD markers. When comparing spatiofrequential models, the best tangent space mapping models significantly outperformed the Euclidean reference models (BrainVol, $p$ = 0.003; HippVol, $p$ = 0.030).\\
Differences between model performances were assessed by statistically comparing squared errors of test set predictions and averaging $p$-values across repeated cross-validation.
Further, we calculated test statistics for evaluating the stand-alone performance of the best models (BrainVol, $p$ = 0.003; Hippvol, $p$ = 0.011).\\
For the prediction of BrainVol (measured during EC resting state) COV yielded lower root-mean-square errors (RMSE) than KEN. Information on the the magnitude of the signal at certain sites -- which is present in the diagonal elements of COV but not KEN matrices -- seem to be essential. Whereas for hippocampus-mediated memory encoding during the WLT, the interaction between brain regions (neuronal networks), as measured by off-diagonal matrix elements, seem to be of predominant importance -- explaining the superior results for KEN. Further, the EEG signal during an active eyes-open task is presumably less normally distributed (even after sophisticated pre-processing) then during a resting EC period, additionally explaining deviating results for COV and KEN.\\ 
Interestingly, TAN$_{log}$ achieved better results than TAN$_{rie}$. Barachant \cite{Barachant2016} also \textit{inter alia} used tangent space mapping with a Log-Euclidean reference point ($M_{G}$) for winning Microsoft's 'mind reading' challenge.
Should future studies support the superiority -- or at least equality -- of TAN$_{log}$, computational cost could be dramatically decreased due to the algorithmic simplicity of Log-Euclidean as compared to Riemannian mean calculation.\\
To the best of our knowledge, this is the first article reporting a Riemannian approach for building QEEG markers of neuronal degeneration. 
\\
\\
\\
\\
\\
\\
\\
\\
\\
\tiny
\begin{table}[]
\caption{Mean, minimum and maximum root-mean-square error (RMSE) of 100 nested cross-validation repetitions for predicting the normalized whole-brain volume (BrainVol) and normalized hippocampus volume (HippVol) for various combinations of measures of dependence (Dep; covariance, COV; Kendall rank correlation, KEN), geometric approaches (Approach; Euclidean, EUC; tangent space mapping with Log-Euclidean mean, TAN$_{log}$, and Riemannian mean, TAN$_{rie}$), and spatial (S), or spatiofrequential (SF) matrix designs (Design).}
  \small
  \label{results}
  \centering
  \begin{tabular}{lllllllll}
    \toprule
 &  \multicolumn{2}{l}{} & \multicolumn{2}{l}{BrainVol} & \multicolumn{2}{r}{HippVol}\\
    \cmidrule{4-9}
 Dep             &  Approach      & Design        & RMSE              & Min                 & Max                 & RMSE        & Min     & Max\\
    \midrule
                        & EUC            & SF            &   1.70E-03        &   1.57E-03          &   2,53E-03         &   2.09E-07          &1.63E-07  &4.26E-07
   \\
    COV                 & TAN$_{log}$   & SF             &\textbf{1.23E-03}  &\textbf{1.10E-03}    &\textbf{1.37E-03}    &   1.86E-07         &1.71E-07  &2.04E-07    \\
                        & TAN$_{rie}$   & SF             &   1.43E-03        &   1.27E-03          &   1,63E-03           &   1.80E-07        &1.67E-07  &2.04E-07
   \\
    \midrule
                        & EUC           & SF             &   1.75E-03        &   1.61E-03          &  2.11E-03        &   1.78E-07             & 1.65E-07 & 2.05E-07
  \\
    KEN                 & TAN$_{log}$   & SF             &   1.56E-03        &   1.43E-03          & 1.77E-03           &   \textbf{1.44E-07}  &\textbf{1.30E-07} & \textbf{1.44E-07}\\
                        & TAN$_{rie}$   & SF             &   1.58E-03        &  1.41E-03           & 1.91E-03           &   1.47E-07           &1.31E-07 & 1.65E-07  
  \\
      \midrule
                        & EUC           & S               &   2.02E-03       & 1.57E-03            & 3.80E-03           &   1.69E-07           &1.55E-07 &2.17E-07
   \\
    COV                 & TAN$_{log}$   & S              &   1.34E-03        & 1.25E-03            &1.50E-03            &   1.77E-07           &1.62E-07 &2.01E-07
   \\
                        & TAN$_{rie}$   & S              &   1.35E-03         &1.24E-03            & 1.46E-03           &   1.75E-07           &1.56E-07 &2.10E-07

   \\
           \midrule
                        & EUC           & S              &   1.73E-03       & 1.59E-03             & 1.90E-03          &  1.68E-07             &1.51E-07 &1.95E-07
  \\
    KEN                 & TAN$_{log}$   & S              &   1.56E-03       & 1.43E-03             &1.79E-03           &  1.79E-07             &1.63E-07 &2.18E-07     
 
  \\
                        & TAN$_{rie}$   & S              &   1.55E-03       &1.44E-03              & 1.73E-03          &  1.81E-07             &1.63E-07 &2.56E-07\\
    \bottomrule
  \end{tabular}
\end{table}
\normalsize
\clearpage 
\bibliography{nips_2017}
\bibliographystyle{authordate1}
\end{document}